\documentclass[sn-basic]{sn-jnl}% Basic Springer Nature Reference Style/Chemistry Reference Style
\usepackage{lineno,hyperref}
\usepackage{amsthm,amsmath,amssymb,mathrsfs}
\usepackage{array}
\usepackage{paralist}
\usepackage{multirow}
\usepackage{booktabs}
\usepackage{makecell}
\jyear{2021}%

\theoremstyle{thmstyleone}%
\theoremstyle{thmstyletwo}%

\theoremstyle{thmstylethree}%

\begin{document}

\title[Learning Rich Features for Gait Recognition]{Learning Rich Features for Gait Recognition by Integrating Skeletons and Silhouettes}


\author[1,2]{\fnm{Yunjie} \sur{Peng}}\email{yunjiepeng@buaa.edu.cn}

\author[2]{\fnm{Kang} \sur{Ma}}\email{kangx.ma@gmail.com}

\author[3]{\fnm{Yang} \sur{Zhang}}\email{zhangyang20@lenovo.com}

\author*[1,3]{\fnm{Zhiqiang} \sur{He}}\email{zqhe1963@gmail.com}

\affil*[1]{\orgdiv{School of Computer Science and Technology}, \orgname{Beihang University}, \orgaddress{\street{Xueyuan Road}, \city{Beijing}, \postcode{100191}, \country{China}}}

\affil[2]{\orgname{Watrix Technology Limited Co. Ltd.}, \orgaddress{\street{XueYuan Road}, \city{Beijing}, \postcode{100083}, \country{China}}}

\affil[3]{\orgname{Lenovo Co. Ltd.}, \orgaddress{\street{Xibeiwang Road}, \city{Beijing}, \postcode{100085}, \country{China}}}


\abstract{
Gait recognition captures gait patterns from the walking sequence of an individual for identification.
Most existing gait recognition methods learn features from silhouettes or skeletons for the robustness to clothing, carrying, and other exterior factors.
The combination of the two data modalities, however, is not fully exploited.
Previous multimodal gait recognition methods mainly employ the skeleton to assist the local feature extraction where the intrinsic discrimination of the skeleton data is ignored.
This paper proposes a simple yet effective Bimodal Fusion (BiFusion) network which mines discriminative gait patterns in skeletons and integrates with silhouette representations to learn rich features for identification.
Particularly, the inherent hierarchical semantics of body joints in a skeleton is leveraged to design a novel Multi-Scale Gait Graph (MSGG) network for the feature extraction of skeletons.
Extensive experiments on CASIA-B and OUMVLP demonstrate both the superiority of the proposed MSGG network in modeling skeletons and the effectiveness of the bimodal fusion for gait recognition.
Under the most challenging condition of walking in different clothes on CASIA-B, our method achieves the rank-1 accuracy of 92.1\%.
}


\maketitle

\section{Introduction}
Gait recognition measures unique physical and behavioral characteristics from the walking pattern of an individual for identification~\cite{ziyuanzhang2019gait}.
Compared with other biometrics such as the face~\cite{yisun2014face}, fingerprint~\cite{maltoni2005fingerprint}, and iris~\cite{wildes1997Iris}, gait can be identified at a distance without the cooperation of target subjects.
It has broad applications in crime prevention, forensic identification, and social security~\cite{larsen2008gait,bouchrika2011gait}.
However, the exterior factors such as clothing, carrying condition, and camera viewpoints greatly change the gait appearance and bring significant challenges to gait recognition.

To alleviate the issue, various methods have been proposed and can be roughly categorized into RGB-based, event-based, silhouette-based, and skeleton-based methods according to the input data modality.
The RGB-based methods~\cite{ziyuanzhang2020gait, li2020gait} usually employ generative networks or skinned multi-person linear (SMPL) models to filter out color and texture information from the raw RGB images for identification.  
These methods are facing the challenge of discarding irrelevant information for gait and require cross-dataset experiments to validate the effectiveness.
The event-based methods~\cite{wang2021eventgait} are proposed with the development of event cameras.
Event cameras capture event streams at each pixel and have the advantages of ultra-low resources consumption and high temporal resolution~\cite{gallego2020event}.
Nevertheless, noisy and asynchronous event streams are relatively hard to deal with.

\begin{figure}  
\centering  
\includegraphics[width=0.99\textwidth]{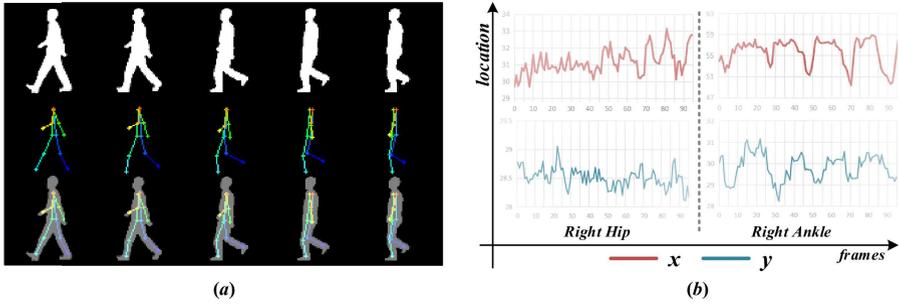}
\caption{(\emph{a}) The visualization of silhouettes and skeletons in a walking sequence;
(\emph{b}) Location changes of the right hip (the joint with larger amplitudes of movement) and the right ankle (the joint with smaller amplitudes of movement) along the \emph{x-axis} and the \emph{y-axis}}\label{fig1}
\end{figure}

The silhouette-based and skeleton-based methods are probably the two most competitive gait recognition methods in the research community.
The silhouette-based methods~\cite{chao2019gaitset,chaofan2020gaitpart,lin2021gaitgl,hou2020gait,xinnan2021gait} extract features from silhouette sequences to eliminate the impact of exterior factors.
These methods are suitable for low-resolution conditions and have achieved state-of-the-art (SOTA) results on public gait datasets.
Despite the superiority, further improvements of these methods are limited: the silhouette only retains the external outline; some body parts information is lost due to the overlapping of limbs and torso during walking, as shown in Fig.~\ref{fig1} (\emph{a}).
The skeleton-based gait recognition methods~\cite{sun2018view,liao2020gait,weizhian2020oumvlp-pose,mao2020gait,teepe2021gaitgraph,xu2021gait} perform human pose estimation~\cite{cao2017openpose,kesun2019pose} first and capture gait patterns from skeleton sequences for recognition.
The skeleton well preserves the body structure information but ignores the discriminative body shape information, resulting in the poor performance of skeleton-based methods.

While the silhouette retains the body shape information and omits some body part clues, the skeleton preserves the body structure information and ignores the shape instead.
The two data modalities are complementary to each other and their combination is expected to be a more comprehensive representation for gait as shown in the third row of Fig.~\ref{fig1} (\emph{a}).
However, previous multimodal gait recognition methods mainly employ the skeleton for local feature extraction on other data modalities, where the intrinsic discrimination of the skeleton data is ignored.
As far as we know, the early work proposed by Boulgouris et al.~\cite{boulgouris2013gait} employs the skeleton structure as a body component locator to assist the local feature extraction of silhouettes.
The global and local silhouette features of automatically labeled components are combined through a pair of Hidden Markov Models for gait recognition.
Similarly, Yao et al. proposed the Collaborative Network~\cite{yao2021gait} which uses the skeleton estimated by HRNet~\cite{kesun2019pose} to locate and extract local features on RGB frames, and employs a transformer encoder to learn the dependence of each correlative local region.

To fully exploit the intrinsic discrimination of the skeleton data and mine complementary clues of the two data modalities, we propose a simple yet effective Bimodal Fusion (BiFusion) network.
Specifically, we adapt the widely-used graph convolution network (GCN) for the graph-structured skeleton data.
The inherent hierarchical semantics of body joints in a skeleton is leveraged to design a novel Multi-Scale Gait Graph (MSGG) network, as well as a task-specific node partitioning strategy, for the feature extraction of skeletons.
The part-based feature-level fusion is applied on features extracted from silhouettes and skeletons to exploit the complementary strengths of the two data modalities for effective gait identification.

In summary, the major contributions of this work lie in three aspects:
\begin{itemize}
\item{We propose a novel Multi-Scale Gait Graph (MSGG) network which automatically captures the gait patterns from the raw skeleton data in a hierarchical manner.}
\item{Different from previous multimodal methods that employ the skeleton for local feature extraction on other data modalities, we design a simple yet effective Bimodal Fusion (BiFusion) network which fully exploits the intrinsic discrimination in skeletons and integrates with silhouette representations to learn rich features for gait recognition.}
\item{Experiments on CASIA-B and OUMVLP demonstrate both the superiority of the proposed MSGG network among skeleton-based methods and the effectiveness of the bimodal fusion in gait identification.}
\end{itemize}

The rest of this paper is organized as follows.
Section 2 gives a summary of related previous works.
Section 3 describes the proposed gait recognition algorithm.
Section 4 presents the implementation details and experimental results.
Section 5 provides our conclusions.

\section{Related work}
\subsection{Silhouette-based Gait Recognition}
\label{related_work_silhouette}
The silhouette-based methods can be roughly divided into three categories according to the way they deal with the temporal information.
The first category~\cite{hanjun2005gait,yu2017gait,himanshu2018gait,lishani2019gait,xu2019gait} regards gait as a template and compresses the silhouettes into an image for identification, e.g. Gait Energy Image (GEI) or Gait Entropy Image (GEnI).
Despite the simplicity, these methods inevitably lose fine-grained temporal and spatial information.
The second~\cite{suibing2018gait,lin2020gait,lin2021gaitgl} regards gait as a video sequence and employs the LSTM-based or 3D-CNN-based model to extract temporal patterns.
These models can capture rich temporal information but require higher computational cost and are relatively hard to train.
The third category~\cite{chao2019gaitset,chaofan2020gaitpart,hou2020gait} regards gait as an unordered set and applies statistical functions to aggregate the temporal information along the set dimension.
It has achieved significant improvements and becomes popular among the research community recently for both simplicity and effectiveness.

In this work, we treat the silhouette sequence as an unordered set and directly employ the GaitPart~\cite{chaofan2020gaitpart} as the silhouette module of the proposed Bimodal Fusion network.
As shown in Fig.~\ref{fig_overview}, the GaitPart module takes a walking silhouette sequence as input and outputs $N$ part features.
In GaitPart, the feature map of a frame is horizontally splitted into multiple parts and each part performs 2D convolution individually to obtain the personalized spatial representation.
After that, the GaitPart calculates the micro-motion representation of each part by aggregating the information from the same parts across adjacent frames (3 or 5 frames), which is implemented as sliding the 1D convolution along the temporal dimension.
The final $N$ part features are then obtained by applying the max pooling operation along the temporal dimension on each micro-motion part features.

\subsection{Skeleton-based Gait Recognition}
Traditional skeleton-based methods~\cite{deng2018gait,liao2020gait,weizhian2020oumvlp-pose} are usually based on handcrafted formulations or rules that extract discriminative parameters from the raw skeleton data for gait recognition.
Deng et al.~\cite{deng2018gait} calculate 4 spatial-temporal parameters (knee stride width, ankle stride width, knee elevation, and ankle elevation) and 4 kinetic parameters (the absolute angle between the vertical direction and the four keypoints, i.e., left knee, left hip, right hip, and right knee) under the Microsoft Kinect coordinates, and employ deterministic learning theory to capture the gait dynamics from the obtained gait parameters for gait identification.
The raw skeleton data predicted by pose estimation methods is not in the form of regular 2D or 3D grids.
Liao et al.~\cite{liao2020gait} define three kinds of spatial-temporal pose features (i.e., joint angle, joint motion, and limb length) based on human prior knowledge and rearrange them to form a feature matrix as the input of a CNN-based model for identification.
These methods rely on handcrafted input features which makes the whole process complex and sub-optimal.
Differently, the recent skeleton-based methods~\cite{mao2020gait,teepe2021gaitgraph,xu2021gait} propose to adapt the GCN for automatically modeling the graph-structured skeleton data.
Benefiting from the powerful graph modeling capability of the GCN, these methods have made significant improvements in skeleton-based gait recognition.

To a certain degree, the skeleton-based GCN methods are closely related to the ST-GCN~\cite{sijieyan2018st-gcn}.
The spatial-temporal graph of ST-GCN is widely adopted due to the ability of modeling variable-length skeleton sequences.
However, the whole architecture of the spatial-temporal graph is inherently flat and the lack of hierarchy structure is especially problematic due to the global pooling operation on the embeddings of all nodes to get an entire graph representation.
Therefore we propose to leverage the inherent hierarchical semantics of body joints in a skeleton and construct a 3-scale hierarchical network named Multi-Scale Gait Graph (MSGG) which aggregates complex multi-scale information for gait recognition.
It is worth noting that the multi-scale GCN is first proposed in CTL~\cite{liu2021videoreid} for video-based person re-identification and the MSGG differs from it in three aspects.
Firstly, the CTL deals with local RGB features extracted (located) by the pose estimation method while the MSGG directly processes the raw skeleton data.
Secondly, the multi-scale skeleton graph defined by CTL is based on human anatomy while that of MSGG is defined by the moving trend of each joint during walking.
Thirdly, the CTL performs multi-layer 3D graph convolution on different body partition scales individually while the MSGG passes information across scales after every spatial-temporal convolution within each scale.

\subsection{Multimodal Biometric Fusion}
Two or more data modalities are employed to extract and combine relevant information for identification in multimodal biometric fusion~\cite{bodla2017fusion,xin2018fusion,dhiman2020multimodalaction,singh2021multimodalaction,singh2021multimodalaction2,dhiman2021multimodalaction}.
Generally, multimodal fusion can be classified into data-level, feature-level, and decision-level fusion according to the level at which the fusion is done.
Compared with data-level and decision-level fusion, feature-level fusion preserves raw information of different modalities and is expected to reach better results~\cite{ross2005fusion,faundez-zanuy2005fusion,shekhar2014fusion}.
In practice, feature-level fusion has been applied in a number of multi-modal biometric identification tasks.
For instance, Bodla et al.\cite{bodla2017fusion} propose a heterogeneous feature fusion network to concatenate both the raw features and the non-linear projected features obtained from two different pre-trained deep CNNs for template-based face recognition.
Another example is that Yang et al.\cite{xin2018fusion} combine face, fingerprint, and finger vein fisher vector features in series and project the fused feature for person identification.

Motivated by the complementary strengths of the silhouettes and skeletons, we adopt feature-level fusion to learn rich features from the two data modalities for effective gait identification.
\begin{figure}  
\centering  
\includegraphics[width=0.99\textwidth]{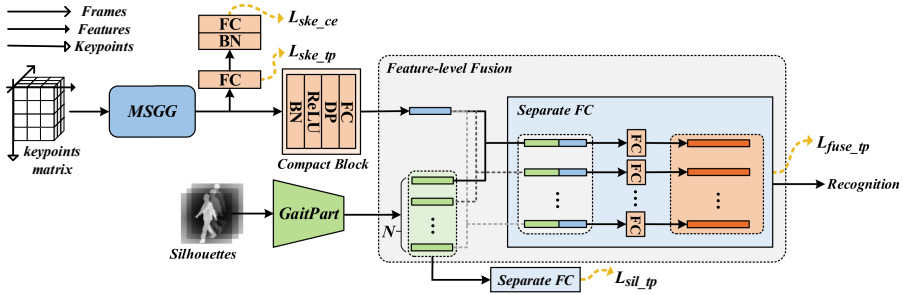}
\caption{Overview of the proposed Bimodal Fusion network. BN for \emph{Batch Normalization}, DP for \emph{Dropout}, and FC for the \emph{Fully Connected Layer}. Separate FC denotes for $N$ FCs performed on $N$ fusion features separately}
\label{fig_overview}  
\end{figure}
\section{Proposed method}
Previous multimodal gait recognition methods~\cite{boulgouris2013gait,yao2021gait} mainly employ the skeleton as a locator to assist the local feature extraction on silhouettes or RGB frames.
The intrinsic discrimination of the skeleton data, however, is remained to be exploited.
Recently, benefiting from the powerful graph modeling capability of the GCN, a lot of methods~\cite{mao2020gait,teepe2021gaitgraph,xu2021gait} have been proposed to capture gait patterns directly from skeleton sequences and have reached pretty good performance on public datasets.
Based on this, we design a novel Multi-Scale Gait Graph (MSGG) network to mine deeper intrinsic discrimination of the skeleton data and attempt to integrate complementary features of silhouettes and skeletons through the proposed Bimodal Fusion (BiFusion) network.

\subsection{Pipeline Overview}
The complementary strengths of the two data modalities, i.e., silhouettes and skeletons, motivate us to fuse them and exploit their relevant information for gait recognition.
The overall architecture of the proposed Bimodal Fusion (BiFusion) network is shown in Fig.~\ref{fig_overview}.
Specifically, we directly employ the GaitPart~\cite{chaofan2020gaitpart} to extract $N$ part features from the silhouette data and design a novel Multi-Scale Gait Graph (MSGG) network to hierarchically aggregate the multi-scale information for the skeleton data.
The part-based feature-level fusion is applied to combine different data modalities for identification.

Formally, given a dataset of $P$ people with identities $p_{i}$, $i\in \{1,2,...,P\}$, we denote the silhouette sequence and the \emph{keypoints matrix}\footnote{The \emph{keypoints matrix} is a 3D matrix that organizes the skeleton sequence data into regular grid formats. Each keypoint in a skeleton contains three initial features, i.e., the \emph{x}, \emph{y} coordinates of the keypoint in the frame and the confidence of the prediction.} generated from a walking sequence of $p_{i}$ as $S_{i}$ and $K_{i}$, respectively.
The proposed network obtains gait features from the bimodal fusion of $S_{i}$ and $K_{i}$ through 3 steps, formulated as:
\begin{equation}
\label{eq_ske}
    k_{i} = CmpB(MSGG(K_{i}))
\end{equation}
\begin{equation}
\label{eq_sil}
    s_{i}^{1},\ s_{i}^{2},...,\ s_{i}^{N} = GaitPart(S_{i})
\end{equation}
\begin{equation}
\label{eq_fuse}
    f\,_{i}^{n} = Fuse_n(s_{i}^{n},k_{i}),\quad n = 1,2,...,N
\end{equation}
where the $MSGG$ generates the skeleton sequence features from the \emph{keypoints matrix} $K_{i}$ and the $GaitPart$ extracts $N$ part-based features from the silhouette sequence $S_{i}$.
Further details about the GaitPart can be found in Sec.~\ref{related_work_silhouette}.
The skeleton-based MSGG module will be introduced in Sec.~\ref{msgg}.
The Compact Block~\cite{hou2020gait} is applied to get the compact features $k_{i}$ for skeletons, abbreviated as $CmpB$ in Eq.~\eqref{eq_ske}.
The structure of the Compact Block is clearly illustrated in Fig.~\ref{fig_overview}.
The function $Fuse_n$ aims to extract specialized information from the compact skeleton sequence features $k_{i}$ for the $n$-th silhouette sequence part features $s_{i}^{n}$ and obtains the proper fusion features $f_{i}^{n}$ for gait identification.
Further details about the $Fuse_n$ are illustrated in Sec.~\ref{bimodal_fusion}.

\subsection{Multi-Scale Gait Graph Network}
\label{msgg}
The proposed Multi-Scale Gait Graph (MSGG) network constructs a pyramid spatial-temporal graph on skeletons with semantics going from shallow to deep.
The information can then be hierarchically aggregated by MSGG for gait recognition.
\begin{figure}  
\centering  
\includegraphics[width=0.99\textwidth]{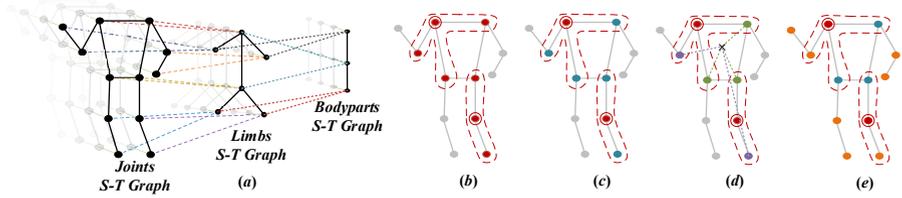}
\caption{(\emph{a}) The proposed pyramid spatial-temporal graph. \emph{S-T} stands for spatial-temporal. (\emph{b}) \emph{Uniform labeling}: all nodes in a neighborhood have the same label. (\emph{c}) \emph{Distance partitioning}: nodes in a neighborhood are divided into two subsets (red and blue) by the distance between the current node and the root node (red). (\emph{d}) \emph{Spatial configuration partitioning}: nodes are labeled into three subsets (red, green, and purple) according to their distances to the gravity center (gray cross) compared with that of the root node (red). (\emph{e}) \emph{Gait temporal partitioning}: nodes are divided into three subsets (red, blue, and orange) where neighbor nodes of the root node (red) are divided into two subsets (blue and orange) according to their moving amplitudes.}\label{fig_pgc} 
\end{figure}

\textbf{Pyramid Spatial-Temporal Graph Construction.}
As shown in Fig.~\ref{fig_pgc} (\emph{a}), the pyramid spatial-temporal graph consists of three subgraphs corresponding to three scales, namely the \emph{joints}, the \emph{limbs}, and the \emph{bodyparts} spatial-temporal graph.
(1) The \emph{joints} spatial-temporal graph is constructed with the joints of skeletons as graph nodes, the natural connectivities in human bodies as spatial edges, and the connections of the same joints across the adjacent frames as temporal edges.
(2) The \emph{limbs} spatial-temporal graph is constructed on the \emph{joints} spatial-temporal graph by pooling every two joints of a limb in the same frame into one node. The reason for this is that the two joints of a limb can represent a physically rigid body in which the same moving trend is shared.
(3) The \emph{bodyparts} spatial-temporal graph is constructed on the \emph{limbs} spatial-temporal graph by pooling every two nodes of a body part in the same frame into one node. The higher-level semantic meaning is reached through grouping the nodes of limbs into corresponding body parts, i.e., torso, arms, and legs.

\textbf{Subgraph Information Aggregation.}
Before diving into the information aggregation on the whole pyramid spatial-temporal graph, we first look at that within each subgraph, i.e.,  the joints, the limbs, or the bodyparts spatial-temporal graph.
Given a skeleton sequence with $T$ frames and $N$ nodes per frame, the subgraph is denoted as $G=(V,E)$.
The node set $V=\{v_{ti} \mid t=1,2,...,T,i=1,2,...,N\}$ and the edge set $E=\{E_{temporal}\,\bigcup\,E_{spatial}\}$, where  $E_{temporal}=\{(v_{ti},v_{t+1i}) \mid t=1,2,...,T-1,\,i=1,2,...,N\}$ contains inter-frame edges which connect the same nodes in consecutive frames and $E_{spatial}=\{(v_{ti},v_{tj}) \mid (i,j)\in H\}$ 
includes intra-frame edges which depict the natural connectivities in a human body based on $H$.
$H$ is a set of connected nodes within a human body defined by the corresponding subgraph in Fig.~\ref{fig_pgc} (\emph{a}).
The information aggregation on $G$ consists of two alternatively performed steps: (\emph{a}) the spatial information aggregation on nodes in a single frame and (\emph{b}) the temporal information aggregation on the same node across the adjacent frames, which are elaborated as the follows.

 (\emph{a}) \emph{Spatial Information Aggregation.} 
We adopt the implementation of the graph convolution in ST-GCN\cite{sijieyan2018st-gcn} to aggregate the spatial information for each frame.
Given the spatial graph $G_t=(V_t, E_{spatial}^{t})$ and the input feature matrix $f_{in}^t$ at the $t$-th frame, the spatial graph convolution is formulated as:
\begin{equation}
\label{eq_sgcn}
    f_{out}^t = \Lambda^{-\frac{1}{2}}(A+I)\Lambda^{\frac{1}{2}}f_{in}^tW
\end{equation}
where $A$ and $I$ are adjacency matrix and identity matrix of $G_t$.
The degree matrix $\Lambda$ is calculated by $\Lambda^{ii}=\sum_j(A^{ij}+I^{ij})$.
Nodes can aggregate information from neighbors with Eq.~\eqref{eq_sgcn}, where $W$ is a trainable weight transform matrix.

In essence, the above formula directly sums the features of a neighbor set $N(v_{ti})=\{v_{tj} \mid d(v_{ti},v_{tj})\leq1\}$, where $d(v_{ti},v_{tj})$ denotes the shortest path from $v_{ti}$ to $v_{tj}$, as the output feature for the node $v_{ti}$.
To improve the expressiveness of the spatial graph convolution, we partition the neighbor set $N(v_{ti})$ of node $v_{ti}$ into a fixed $K$ subsets and propagate the information for each subset separately.
Here we define the mapping $l_{ti}: N(v_{ti})\rightarrow \{0,1,...,K-1\}$ which maps a node in the neighbor set $N(v_{ti})$ to its subset label.
The spatial graph convolution with $K$ subsets is formulated as:
\begin{equation}
\label{eq_sgcn_k}
    f_{out}^t = \sum\limits_{k=1}^K\Lambda_k^{-\frac{1}{2}}(A_k+I)\Lambda_k^{\frac{1}{2}}f_{in}^tW_k
\end{equation}
where $\Lambda_k^{-\frac{1}{2}}(A_k+I)\Lambda_k^{\frac{1}{2}}$ denotes the normalized adjacency matrix and $W_k$ denotes a trainable weight transform matrix for the $k$-th subset of each neighbor set on graph $G_t$, respectively.

(\emph{a1}) \emph{Partition Strategy.}
Given the formulation of graph convolution in Eq.~\eqref{eq_sgcn_k}, it is important to design a proper node partition strategy to implement the mapping function $l_{ti}$.
As shown in Fig.~\ref{fig1} (\emph{b}), the joints with larger amplitudes of movement are more likely to retain the gait periodicity than that with smaller moving amplitudes from the analysis of the temporal changes in skeleton data.
Therefore, we propose to partition the neighbor set $N(v_{ti})$ based on the moving amplitudes of each neighbor node $v_{tj}$, formulated as:
\begin{equation}
{l_{ti}(v_{tj})}=\begin{cases}
0 ,&{\text{if}}\ i=j;\\
1 ,&{\text{if}}\ v_{tj}\in Positive;\\
2 ,&{\text{if}}\ v_{tj}\in Negative.
\end{cases}
\end{equation}
where the nodes with larger amplitudes of movement are regarded as \emph{Positive} and the nodes with smaller moving amplitudes are regarded as \emph{Negative}\footnote{\emph{Positive}: right elbow, right knee, left elbow, left knee, right wrist, right ankle, left wrist, and left ankle. \emph{Negative}: right shoulder, right hip, left shoulder, left hip. \emph{Positive} and \emph{Negative} nodes of the limbs spatial-temporal graph and the bodyparts spatial-temporal graph are similarly defined.}, except for the current node $v_{ti}$.
We name this strategy as \emph{Gait temporal partitioning} and demonstrate its superiority in Sec.~\ref{ablation_msgg} through the comparison with different partition strategies proposed in the ST-GCN~\cite{sijieyan2018st-gcn}.
Different strategies are clearly illustrated in Fig.~\ref{fig_pgc} (\emph{b})-(\emph{e}).

(\emph{a2}) \emph{Edge Importance Weighting.}
To find out the contribution of different nodes in the $k$-th subset of the neighbor set $N(v_{ti})$, we add a learnable edge importance weight matrix $W_E$ for the spatial graph convolution.
It is straightforward to implement the edge importance weighting by substituting the matrix $(A_k+I)$ in Eq.~\eqref{eq_sgcn_k} with $(A_k+I)\bigotimes W_E$, where $\bigotimes$ denotes the Hadamard product.

(\emph{b}) \emph{Temporal Information Aggregation.}
% In the construction of the spatial-temporal graph for a skeleton sequence, the temporal changes of each node are passed through the connections of the same node across consecutive frames.
%
The well-ordered temporal axis of each node enables the MSGG to simply aggregate the temporal information within adjacent $\Gamma$ frames rooted at $v_{ti}$ as:
\begin{equation}
\label{eq_tagg}
f_{out}^{t,i}=
    \begin{bmatrix}
    f_{in}^{t-\left \lfloor \Gamma/2 \right \rfloor,i},\cdots,f_{in}^{t,i},\cdots,f_{in}^{t+\left \lfloor \Gamma/2 \right \rfloor,i}
    \end{bmatrix}
    W_{\Gamma\times 1}^{i}
\end{equation}
where $f_{in}^{t,i}$ denotes the input feature vector of node $v_{ti}$.
The learnable weight matrix $W_{\Gamma\times 1}^{i}$ is employed to aggregate the temporal information of adjacent $\Gamma$ frames for node $v_{ti}$.
\begin{figure}  
\centering  
\includegraphics[width=0.99\textwidth]{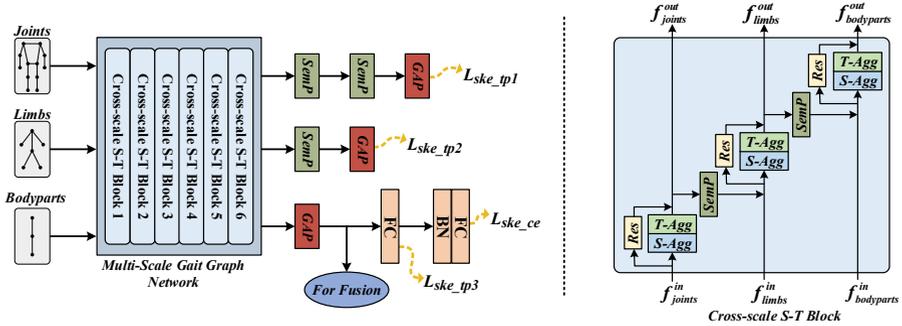}
\caption{The overall architecture of the proposed \emph{muilti-scale gait graph (MSGG)} network. \emph{S-T} is short for spatial-temporal and \emph{SemP} is short for semantic pooling. \emph{GAP} denotes for the global pooling operation on the embeddings of all nodes. BN for \emph{Batch Normalization} and FC for the \emph{Fully Connected Layer}. \emph{S-Agg} denotes for the spatial information aggregation defined by Eq.~\eqref{eq_sgcn_k} and \emph{T-Agg} denotes for the temporal information aggregation defined by Eq.~\eqref{eq_tagg}. \emph{Res} for the residual connection}\label{fig_msgg}
\end{figure}

\textbf{Multi-Scale Gait Graph Network.}
The proposed Multi-Scale Gait Graph (MSGG) network is built on the pyramid spatial-temporal graph constructed in Sec.~\ref{msgg}.
Benefitting from the natural hierarchical structure defined by semantics, MSGG is able to exploit diverse multi-scale features and aggregate the information in a hierarchical way.
As shown in Fig.~\ref{fig_msgg}, the proposed MSGG consists of 3 branches that are conducted on the joints, the limbs, and the bodyparts spatial-temporal subgraph respectively.
In MSGG, there are 6 Cross-scale Spatial-Temporal (S-T) Blocks, and each block completes the spatial-temporal information aggregation within each subgraph and the message passing across subgraphs.

The detailed structure of the Cross-scale S-T Block is shown in the right half of Fig.~\ref{fig_msgg}.
The spatial-temporal information aggregation on each subgraph within a Cross-scale S-T Block is implemented by the execution of a spatial information aggregation (Eq.~\eqref{eq_sgcn_k}) and a temporal information aggregation (Eq.~\eqref{eq_tagg}). 
To alleviate the problem of gradient vanishing, a residual connection is added to the spatial-temporal information aggregation except for the first Cross-scale S-T Block.
Every time the spatial-temporal information aggregation has been done, the Cross-scale S-T Block will pass messages from the graph with lower semantic-level to the adjacent graph with higher semantic-level, i.e., from the joints subgraph to the limbs subgraph and from the limbs subgraph to the bodyparts subgraph, denoted as the \emph{SemP} in Fig.~\ref{fig_msgg}.

\emph{Semantic Pooling.}
We name the message passing between adjacent subgraphs as the \emph{Semantic Pooling (SemP)}.
The goal of the \emph{SemP} is to deliver the information from the graph with lower semantics to the graph with higher semantics.
Given the mapping $M$ between adjacent subgraphs, the message passing is formulated as:
\begin{equation}
    f_{out}^{ti} = f_{in}^{ti} + SemP(f_{in}^{ta'}, f_{in}^{tb'}),\ ((a,b)\rightarrow i)\in M
\end{equation}
where $f_{in}^{ta'}$ and $f_{in}^{tb'}$ denote the feature vector of $v_{ta}$ and $v_{tb}$ in the joints (limbs) subgraph, and $f_{in}^{ti}$ denotes the feature vector of $v_{ti}$ in the limbs (bodyparts) subgraph.
Element $((a,b)\rightarrow i)$ in $M$ denotes that the node $v_{ta}$ and the node $v_{tb}$ in a lower semantic graph are defined to be mapped to the node $v_{ti}$ in the current graph.
The mapping $M$ between adjacent subgraphs is clearly illustrated in Fig.~\ref{fig_pgc} (\emph{a}).
For simplicity, we use the $average(\cdot)$ function as the implementation of \emph{SemP}.

\subsection{Bimodal Fusion Network}
\label{bimodal_fusion}
The proposed Bimodal Fusion (BiFusion) network exploits complementary strengths of the two data modalities, i.e., silhouettes and skeletons, for gait recognition.
As shown in Fig.~\ref{fig_overview}, given $N$ silhouette part features generated from the GaitPart module and the compact skeleton features extracted from the MSGG module followed by a Compact Block, the part-based bimodal fusion (the function $Fuse_n$ in Eq.~\eqref{eq_fuse}) is implemented with the feature concatenation followed by a fully connected layer.
We concatenate the skeleton sequence features reduced by the Compact Block~\cite{hou2020gait} after each part-based silhouette sequence features and employ the fully connected layer (FC) to get the bimodal fusion for each part.
This way of feature-level fusion is expected to extract specialized information from the skeleton data modality for the corresponding part features captured by the silhouette data modality.
Therefore, a more comprehensive representation for gait recognition is obtained.

\section{Experiments}
The skeleton data is required for the proposed Bimodal Fusion (BiFusion) network.
Currently, most cross-view large open-source gait datasets such as OU-LP~\cite{iwama2012isir} and OU-ISIR Treadmill~\cite{makihara2012isirtreadmill} do not provide the skeleton data or give the access to raw RGB frames.
In this case, empirical experiments are conducted on two publicly available gait datasets: CASIA-B and OUMVLP.
Both datasets are labeled with person identities.
For each dataset, we evaluate the performance of MSGG among skeleton-based methods and compare the BiFusion network with previous state-of-the-art methods.
Further ablation experiments are conducted on CASIA-B, which contains the skeleton data of high quality and considers about walkings with varieties.

\subsection{Datasets}
\label{dataset}
\textbf{CASIA-B}~\cite{shiqiyu2006gait} is a popular gait dataset widely used in the research community.
It contains 124 subjects where each subject has 10 walkings under three different conditions, i.e., 6 walkings in normal (NM), 2 walkings with a bag (BG), and 2 walkings in different clothes (CL).
Each walking is recorded by 11 cameras located at different views simultaneously. 
The views for setting 11 cameras are uniformly distributed in $[0^{\circ},180^{\circ}]$ at an interval of $18^{\circ}$.
In total, there are $(6+2+2)\times 11=110$ walking sequences per subject.
Our experiments take the first 74 subjects as the training set and the rest 50 as the testing set.
For evaluation, the first 4 normal walking sequences of each subject are regarded as the gallery and the rest are regarded as the probe.

\noindent\textbf{OUMVLP}~\cite{takemura2018oumvlp} is a large public gait dataset that released only the silhouette data at the beginning.
Recently, the skeleton data of OUMVLP estimated by AlphaPose~\cite{fang2017alphapose} and Openpose~\cite{cao2017openpose} have been released by An et al.~\cite{weizhian2020oumvlp-pose}, named as the OUMVLP-Pose dataset.
Therefore, the OUMVLP is an open-source dataset that has released both the silhouette data~\cite{takemura2018oumvlp} and the skeleton data~\cite{weizhian2020oumvlp-pose}.
It contains 10307 subjects, 14 views ($[0^{\circ},90^{\circ}]$ and $[180^{\circ},270^{\circ}]$ at an interval of $15^{\circ}$) per subject, and 2 walking sequences (\#00-01) per view.
The division of the training set and the testing set provided in OUMVLP~\cite{takemura2018oumvlp} is quite different from that in OUMVLP-Pose~\cite{weizhian2020oumvlp-pose}.
For the comparison with previous SOTA methods, main experiments are conducted under the division provided in OUMVLP~\cite{takemura2018oumvlp}. 
The 10307 subjects are divided into two disjoint groups: 5153 training and 5154 testing subjects.
For evaluation, sequences \#01 are kept in the gallery and sequences \#00 are regarded as the probe.

\subsection{Implementation Details}
\label{implementation_details}
All models are implemented with PyTorch~\cite{paszke2019pytorch} and trained with NVIDIA Titan-V GPUs.

\textbf{Input}
The silhouettes are pre-processed into the resolution of $64\times64$ using the methods described in~\cite{chao2019gaitset}.
Both the Openpose~\cite{cao2017openpose} and the HRNet~\cite{kesun2019pose} are applied in CASIA-B for pose estimation.
For OUMVLP-Pose~\cite{weizhian2020oumvlp-pose}, we conduct experiments on the released AlphaPose dataset.
As shown in Fig.~\ref{fig_pgc} (\emph{a}), 12 keypoints of the limbs and torso are used in this work.
The number of subjects, the number of sequences per subject, and the number of frames per sequence in a mini-batch are set to $(8,16,30)$ for CASIA-B, $(32,16,30)$ for OUMVLP silhouettes, and $(32,16,18)$ for OUMVLP skeletons.

\textbf{Network}
We use the settings in GaitPart~\cite{chaofan2020gaitpart} for the silhouette module of our Bimodal Fusion network.
There are 6 Cross-scale Spatial-Temporal Blocks in the proposed Multi-Scale Gait Graph (MSGG) network.
We set the number of channels in the first two blocks, the second two blocks, and the third two blocks as $(16,32,64)$ for CASIA-B and $(32,64,128)$ for OUMVLP.
The dropping ratio of the $Dropout$ layer and the output dimension of the $FC$ layer in the Compact Block are set to $(0.3,32)$ for CASIA-B and $(0.65,32)$ for OUMVLP.
The output dimension of each fused part features is set to 128 for CASIA-B and 256 for OUMVLP.

\textbf{Loss}
The cross-entropy loss and Batch All $(BA_+)$ triplet loss~\cite{hermans2017tripletloss} are applied to train the network.
The margin threshold $m$ for triplet loss is set to 0.2.
The triplet loss $L_{sil\_tp}$ is employed for pretraining the GaitPart module.
And the loss $L_{MSGG}$ for pretraining the MSGG module is computed as:
\begin{equation}
    L_{MSGG} = \alpha L_{ske\_tp1} + \beta L_{ske\_tp2} + \gamma L_{ske\_tp} + L_{ske\_ce}
\end{equation}
where $L_{ske\_tp1}$, $L_{ske\_tp2}$, and $L_{ske\_tp}$ are triplet losses calculated at the \emph{joints}, the \emph{limbs}, and the \emph{bodyparts} branches respectively.
$L_{ske\_ce}$ denotes the cross-entropy loss added after $FC$ at the \emph{bodyparts} branch.
The $\alpha$, $\beta$, and $\gamma$ are loss weights of the joints, the limbs, and the bodyparts layer respectively.
The $\alpha$, $\beta$, and $\gamma$ are set to 3, 2, and 1 so that the model gives top priority to the underlying information.
The total loss for the global training is computed as:
\begin{equation}
    L = L_{sil\_tp} + L_{ske\_tp} + L_{ske\_ce}
\end{equation}

\textbf{Optimizer}
The silhouette-based GaitPart module is pretrained as in~\cite{chaofan2020gaitpart}.
The SGD optimizer with momentum is adopted for MSGG pretraining and global training.
In the pretraining of the MSGG, the learning rate is set to 0.1 and will be scaled to its 1/10 per 25K iterations for CASIA-B (per 75K iterations for OUMVLP) three times until convergence.
In global training, the learning rate for the GaitPart module and the MSGG module is set to 1e-4, and for the rest is set to 0.1.
The learning rate will be scaled to its 1/10 per 2K iterations for CASIA-B (per 4K iterations for OUMVLP) four times until convergence.
We use $momentum=0.9$ and $weight\ decay=5e$-$4$ for the optimization.
It is worth noting that all the experiments in this paper did not use any additional data other than the current dataset.

\textbf{Testing}
Given a query $Q$, the goal is to retrieve all the samples with the same identity in gallery set $\mathbb{G}$.
The $Q$ is fed into the proposed BiFusion network to generate multiple part-based features $\{F_Q^1,F_Q^2,...,F_Q^N\}$.
And the same process is applied on each sample $G$ in gallery $\mathbb{G}$ to obtain multiple part-based features $\{F_G^1,F_G^2,...,F_G^N\}$.
We measure the similarity between $Q$ and $G$ by calculating the averaged Euclidean distance as $\frac{1}{N}\sum_{n=1}^{N}\left \| F_Q^n-F_G^n \right \|$.
\begin{table}[!t]
\begin{center}
\newcommand{\bftab}[1]{{\fontseries{b}\selectfont#1}}
\caption{The rank-1 accuracy (\%) on CAISA-B across different views, excluding the identical-view cases. Based on the walking condition, probe sequences are grouped into three subsets, i.e., NM, BG, and CL. The 3D Pose~\cite{chen20173dpose} takes the 2D poses estimated by Openpose~\cite{cao2017openpose} as input. BiFusion stands for the proposed Bimodal Fusion network\label{table_casia-b}}
\resizebox{0.99\textwidth}{0.37\textheight}{
\begin{tabular}{ccccccccccccccc}
\toprule
\multicolumn{2}{c}{Gallery NM} & 
\multicolumn{11}{c}{$0^{\circ}-180^{\circ}$} & 
\multirow{2}{*}{mean} \\ \cline{1-13}

\multicolumn{1}{c}{Probe} & 
\multicolumn{1}{c}{Methods} &
 $0^{\circ}$ & $18^{\circ}$ & $36^{\circ}$ & $54^{\circ}$ & $72^{\circ}$ & $90^{\circ}$ & $108^{\circ}$ & $126^{\circ}$ & $144^{\circ}$ & $162^{\circ}$ &
\multicolumn{1}{l}{$180^{\circ}$} &\\ \cline{1-14}

\multicolumn{1}{c}{\multirow{14}{*}{NM}} &
\multicolumn{1}{c}{PoseGait~\cite{liao2020gait} (3D Pose)} & 
55.3 &69.6 &73.9 &75.0 &68.0 &68.2 &71.1 &72.9 &76.1 &70.4 &
\multicolumn{1}{l}{55.4} & \multicolumn{1}{c}{68.7}\\ 

\multicolumn{1}{c}{} &
\multicolumn{1}{c}{ST-GCN~\cite{sijieyan2018st-gcn} (HRNet)} & 
86.6 &87.5 &85.7 &88.6 &81.2 &82.8 &82.2 &86.0 &87.2 &90.3 &
\multicolumn{1}{l}{81.4} & \multicolumn{1}{c}{85.4}\\ 

\multicolumn{1}{c}{} &
\multicolumn{1}{c}{GaitGraph~\cite{teepe2021gaitgraph} (HRNet)} & 
85.3 &88.5 &91.0 &92.5 &87.2 &86.5 &88.4 &89.2 &87.9 &85.9 &
\multicolumn{1}{l}{81.9} & \multicolumn{1}{c}{87.7}\\ 

\multicolumn{1}{c}{} &
\multicolumn{1}{c}{MSGG(ours, Openpose)} &
74.3 &84.3 &85.2 &85.8 &85.2 &83.2 &86.4 &87.3 &87.7 &84.6 &
\multicolumn{1}{l}{81.9} & \multicolumn{1}{c}{84.2}\\ 

\multicolumn{1}{c}{} &
\multicolumn{1}{c}{MSGG(ours, \quad HRNet)} &
\bftab{88.8} &\bftab{92.6} &\bftab{94.2} &\bftab{94.0} &\bftab{93.0} &\bftab{93.9} &\bftab{92.3} &\bftab{94.5} &\bftab{94.4} &\bftab{94.9} &
\multicolumn{1}{l}{\bftab{90.9}} &
\multicolumn{1}{c}{\bftab{93.0}}\\ \cline{2-14}

\multicolumn{1}{c}{} &
\multicolumn{1}{c}{CNN-Ensemble~\cite{zifengwu2017gait}} &
88.7 &95.1 &98.2 &96.4 &94.1 &91.5 &93.9 &97.5 &98.4 &95.8 & 
\multicolumn{1}{c}{85.6} & \multicolumn{1}{c}{94.1}\\ 

\multicolumn{1}{c}{} &
\multicolumn{1}{c}{GaitNet~\cite{ziyuanzhang2020gait}} &
93.1 &92.6 &90.8 &92.4 &87.6 &95.1 &94.2 &95.8 &92.6 &90.4 &
\multicolumn{1}{c}{90.2} & \multicolumn{1}{c}{92.3}\\ 

\multicolumn{1}{c}{} &
\multicolumn{1}{c}{GaitSet~\cite{chao2019gaitset}} &
90.8 &97.9 & 99.4 &96.9 &93.6 &91.7 &95.0 &97.8 &98.9 &96.8 & 
\multicolumn{1}{c}{85.8} & \multicolumn{1}{c}{95.0}\\ 

\multicolumn{1}{c}{} &
\multicolumn{1}{c}{GaitPart~\cite{chaofan2020gaitpart}} &
94.1 &98.6 &99.3 &98.5 &94.0 &92.3 &95.9 &98.4 &99.2 &97.8 & 
\multicolumn{1}{c}{90.4} & \multicolumn{1}{c}{96.2}\\ 

\multicolumn{1}{c}{} &
\multicolumn{1}{c}{MvGGAN~\cite{xinchen2021gait}} &
94.8 &99.0 &\bftab{99.7} &99.2 &96.6 &93.7 &96.3 &98.6 &99.2 &98.2 & 
\multicolumn{1}{c}{92.3} & \multicolumn{1}{c}{97.1}\\ 

\multicolumn{1}{c}{} &
\multicolumn{1}{c}{GaitGL~\cite{lin2021gaitgl}} &
96.0 &98.3 &99.0 &97.9 &96.9 &95.4 &97.0 &98.9 &99.3 &98.8 &
\multicolumn{1}{c}{94.0} & \multicolumn{1}{c}{97.4}\\ \cline{2-14}

\multicolumn{1}{c}{} &
\multicolumn{1}{c}{silhouette-module(base)} &
93.4 &98.0 &99.1 &97.5 &94.1 &91.9 &96.4 &97.6 &99.0 &97.9 &
\multicolumn{1}{c}{91.9} & \multicolumn{1}{c}{96.1}\\

\multicolumn{1}{c}{} &
\multicolumn{1}{c}{BiFusion(ours, Openpose)} &
97.6 &\bftab{99.2} &99.2 &99.0 &97.7 &95.8 &97.9 &98.6 &99.2 &\bftab{99.6} &
\multicolumn{1}{c}{96.2} & \multicolumn{1}{c}{98.2}\\

\multicolumn{1}{c}{} &
\multicolumn{1}{c}{BiFusion(ours, \quad HRNet)} &
\bftab{98.0} &99.1 &99.5 &\bftab{99.3} &\bftab{98.7} &\bftab{97.5} &\bftab{98.5} &\bftab{99.1} &\bftab{99.6} &99.5 &
\multicolumn{1}{c}{\bftab{96.8}} & \multicolumn{1}{c}{\bftab{98.7}}\\ \cline{1-14}

\multicolumn{1}{l}{\multirow{14}{*}{BG}} &
\multicolumn{1}{c}{PoseGait~\cite{liao2020gait} (3D Pose)} & 
35.3 &47.2 &52.4 &46.9 &45.5 &43.9 &46.1 &48.1 &49.4 &43.6 &
\multicolumn{1}{l}{31.1} & \multicolumn{1}{c}{44.5}\\ 

\multicolumn{1}{c}{} &
\multicolumn{1}{c}{ST-GCN~\cite{sijieyan2018st-gcn} (HRNet)} & 
68.0 &68.8 &62.8 &64.0 &64.2 &60.9 &60.9 &63.2 &64.3 &64.7 &
\multicolumn{1}{l}{51.1} & \multicolumn{1}{c}{63.0}\\ 

\multicolumn{1}{c}{} &
\multicolumn{1}{c}{GaitGraph~\cite{teepe2021gaitgraph} (HRNet)} & 
75.8 &76.7 &75.9 &76.1 &71.4 &\bftab{73.9} &\bftab{78.0} &74.7 &75.4 &75.4 &
\multicolumn{1}{l}{69.2} & \multicolumn{1}{c}{74.8}\\ 

\multicolumn{1}{c}{} &
\multicolumn{1}{c}{MSGG(ours, Openpose)} &
59.9 &65.8 &70.3 &69.2 &67.6 &66.8 &65.9 &68.0 &69.8 &66.8 &
\multicolumn{1}{l}{63.0} & \multicolumn{1}{c}{66.6}\\ 

\multicolumn{1}{c}{} & 
\multicolumn{1}{c}{MSGG(ours, \quad HRNet)}  &
\bftab{77.9} &\bftab{81.3} &\bftab{81.7} &\bftab{80.2} &\bftab{78.2} &73.8 &76.5 &\bftab{77.0} &\bftab{78.6} &\bftab{80.5} &
\multicolumn{1}{l}{\bftab{73.0}} & \multicolumn{1}{c}{\bftab{78.1}}\\ \cline{2-14}

\multicolumn{1}{c}{} & 
\multicolumn{1}{c}{CNN-LB~\cite{zifengwu2017gait}} &
64.2 &80.6 &82.7 &76.9 &64.8 &63.1 &68.0 &76.9 &82.2 &75.4 &
\multicolumn{1}{c}{61.3} & \multicolumn{1}{c}{72.4}\\ 

\multicolumn{1}{c}{} &
\multicolumn{1}{c}{GaitNet~\cite{ziyuanzhang2020gait}} &
88.8 &88.7 &88.7 &94.3 &85.4 &\bftab{92.7} &91.1 &92.6 &84.9 &84.4 &
\multicolumn{1}{c}{86.7} & \multicolumn{1}{c}{88.9}\\ 

\multicolumn{1}{c}{} &
\multicolumn{1}{c}{GaitSet~\cite{chao2019gaitset}} &
83.8 &91.2 &91.8 &88.8 &83.3 &81.0 &84.1 &90.0 &92.2 &94.4 & 
\multicolumn{1}{c}{79.0} & \multicolumn{1}{c}{87.2}\\ 

\multicolumn{1}{c}{} &
\multicolumn{1}{c}{GaitPart~\cite{chaofan2020gaitpart}} &
89.1 &94.8 &96.7 &95.1 &88.3 &84.9 &89.0 &93.5 &96.1 &93.8 & 
\multicolumn{1}{c}{85.8} & \multicolumn{1}{c}{91.5}\\ 

\multicolumn{1}{c}{} &
\multicolumn{1}{c}{MvGGAN~\cite{xinchen2021gait}} &
92.4 &94.7 &97.2 &94.6 &88.7 &83.6 &87.8 &93.8 &96.3 &95.2 & 
\multicolumn{1}{c}{86.8} & \multicolumn{1}{c}{91.9}\\ 

\multicolumn{1}{c}{} &
\multicolumn{1}{c}{GaitGL~\cite{lin2021gaitgl}} &
92.6 &96.6 &96.8 &95.5 &93.5 &89.3 &92.2 &96.5 &98.2 &96.9 &
\multicolumn{1}{c}{91.5} & \multicolumn{1}{c}{94.5}\\ \cline{2-14}

\multicolumn{1}{c}{} &
\multicolumn{1}{c}{silhouette-module(base)} &
90.1 &95.8 &96.0 &93.6 &88.0 &84.4 &88.1 &93.4 &96.0 &93.7 &
\multicolumn{1}{c}{86.0} & \multicolumn{1}{c}{91.4}\\

\multicolumn{1}{c}{} &
\multicolumn{1}{c}{BiFusion(ours, Openpose)} &
\bftab{96.3} &\bftab{98.0} &97.8 &96.6 &93.6 &90.0 &93.4 &96.0 &98.1 &\bftab{98.6} &
\multicolumn{1}{c}{92.0} & \multicolumn{1}{c}{95.5}\\

\multicolumn{1}{c}{} &
\multicolumn{1}{c}{BiFusion(ours, \quad HRNet)} &
95.8 &97.9 &\bftab{98.2} &\bftab{97.6} &\bftab{94.4} &91.6 &\bftab{93.9} &\bftab{96.6} &\bftab{98.5} &98.3 &
\multicolumn{1}{c}{\bftab{93.1}} & \multicolumn{1}{c}{\bftab{96.0}}\\ \cline{1-14}

\multicolumn{1}{l}{\multirow{14}{*}{CL}} & 
\multicolumn{1}{c}{PoseGait~\cite{liao2020gait} (3D Pose)} & 
24.3 &29.7 &41.3 &38.8 &38.2 &38.5 &41.6 &44.9 &42.2 &33.4 &
\multicolumn{1}{l}{22.5} & \multicolumn{1}{c}{36.0}\\

\multicolumn{1}{c}{} &
\multicolumn{1}{c}{ST-GCN~\cite{sijieyan2018st-gcn} (HRNet)} & 
59.3 &60.8 &55.3 &58.5 &55.5 &54.6 &56.2 &57.5 &62.0 &65.3 &
\multicolumn{1}{l}{60.1} & \multicolumn{1}{c}{58.6}\\ 

\multicolumn{1}{c}{} &
\multicolumn{1}{c}{GaitGraph~\cite{teepe2021gaitgraph} (HRNet)} & 
\bftab{69.6} &66.1 &\bftab{68.8} &67.2 &64.5 &62.0 &\bftab{69.5} &65.6 &65.7 &66.1 &
\multicolumn{1}{l}{64.3} & \multicolumn{1}{c}{66.3}\\ 

\multicolumn{1}{c}{} &
\multicolumn{1}{c}{MSGG(ours, Openpose)} &
46.5 &54.9 &55.2 &54.7 &54.7 &57.3 &57.3 &58.3 &61.3 &59.6 & \multicolumn{1}{l}{54.6} & \multicolumn{1}{c}{55.9}\\ 

\multicolumn{1}{c}{} & 
\multicolumn{1}{c}{MSGG(ours, \quad HRNet)} &
62.2 &\bftab{67.4} &66.2 &\bftab{70.2} &\bftab{68.8} &\bftab{66.2} &67.4 &\bftab{69.2} &\bftab{71.1} &\bftab{73.4} &
\multicolumn{1}{l}{\bftab{69.7}} & \multicolumn{1}{c}{\bftab{68.3}}\\ \cline{2-14}

\multicolumn{1}{c}{} &
\multicolumn{1}{c}{CNN-LB~\cite{zifengwu2017gait}} &
37.7 &57.2 &66.6 &61.1 &55.2 &54.6 &55.2 &59.1 &58.9 &48.8 &
\multicolumn{1}{c}{39.4} & \multicolumn{1}{c}{54.0}\\ 

\multicolumn{1}{c}{} &
\multicolumn{1}{c}{GaitNet~\cite{ziyuanzhang2020gait}} &
50.1 &60.7 &72.4 &72.1 &74.6 &78.4 &70.3 &68.2 &53.5 &44.1 & 
\multicolumn{1}{c}{40.8}& \multicolumn{1}{c}{62.3}\\ 

\multicolumn{1}{c}{} &
\multicolumn{1}{c}{GaitSet~\cite{chao2019gaitset}} &
61.4 &75.4 &80.7 &77.3 &72.1 &70.1 &71.5 &73.5 &73.5 &68.4 & 
\multicolumn{1}{c}{50.0} & \multicolumn{1}{c}{70.4}\\ 

\multicolumn{1}{c}{} &
\multicolumn{1}{c}{GaitPart~\cite{chaofan2020gaitpart}} &
70.7 &85.5 &86.9 &83.3 &77.1 &72.5 &76.9 &82.2 &83.8 &80.2 & 
\multicolumn{1}{c}{66.5} & \multicolumn{1}{c}{78.7}\\

\multicolumn{1}{c}{} &
\multicolumn{1}{c}{MvGGAN~\cite{xinchen2021gait}} &
70.5 &77.9 &82.5 &82.7 &77.4 &73.6 &73.8 &77.8 &77.6 &72.5 & 
\multicolumn{1}{c}{64.8} & \multicolumn{1}{c}{75.6}\\

\multicolumn{1}{c}{} &
\multicolumn{1}{c}{GaitGL~\cite{lin2021gaitgl}} &
76.6 &90.0 &90.3 &87.1 &84.5 &79.0 &84.1 &87.0 &87.3 &84.4 &
\multicolumn{1}{c}{69.5} & \multicolumn{1}{c}{83.6}\\ \cline{2-14}

\multicolumn{1}{c}{} &
\multicolumn{1}{c}{silhouette-module(base)} &
75.1 &84.9 &87.4 &82.3 &77.2 &74.8 &79.6 &83.8 &83.7 &80.3 &
\multicolumn{1}{c}{65.8} & \multicolumn{1}{c}{79.5}\\

\multicolumn{1}{c}{} &
\multicolumn{1}{c}{BiFusion(ours, Openpose)} &
85.1 &92.6 &93.8 &92.0 &87.8 &86.5 &89.8 &91.6 &92.0 &92.0 &
\multicolumn{1}{c}{82.5} & \multicolumn{1}{c}{89.6}\\

\multicolumn{1}{c}{} &
\multicolumn{1}{c}{BiFusion(ours, \quad HRNet)} &
\bftab{88.7} &\bftab{93.9} &\bftab{95.6} &\bftab{93.8} &\bftab{91.4} &\bftab{89.4} &\bftab{92.3} &\bftab{93.8} &\bftab{94.2} &\bftab{93.7} &
\multicolumn{1}{c}{\bftab{86.2}} & \multicolumn{1}{c}{\bftab{92.1}}\\ \bottomrule
\end{tabular}
}
\end{center}
\end{table}

\subsection{Main Results}
\textbf{CASIA-B}
In the testing phase of CASIA-B, the probe sequences are grouped into three subsets, i.e., NM, BG, and CL.
The evaluation is performed on each subset separately.
All the results are averaged on gallery views excluding the identical-view cases.

Table \ref{table_casia-b} shows the performance comparison on CASIA-B for both the MSGG module and the BiFusion network.
As listed in the first block of different probes, we compare the MSGG with PoseGait~\cite{liao2020gait} and GaitGraph~\cite{teepe2021gaitgraph}, where only the estimated skeleton data is taken as input.
Although the GaitGraph performs better at few views in BG and CL, mean accuracies at different probes show the superiority of the proposed MSGG module over other skeleton-based methods under the same 2D pose estimation method (3D Pose used in PoseGait is based on Openpose).
Moreover, with the high-quality skeleton data provided by HRNet~\cite{kesun2019pose}, MSGG obtains impressive results (NM-93.0\%, BG-78.1\%, CL-68.3\%) on CASIA-B for skeleton-based gait recognition.
The second and the third blocks of different probes in Table \ref{table_casia-b} show the performance comparison on CASIA-B for our BiFusion network.
Taking advantage of the well-preserved body structure information in skeleton data, the BiFusion network reaches new state-of-the-art performance on CASIA-B (NM-98.7\%, BG-96.0\%, CL-92.1\%).
Under the most challenging condition of walking in different clothes, the proposed method improves the rank-1 accuracy on GaitGL~\cite{lin2021gaitgl} by 8.5\%.
This improvement shows the great potential of utilizing the natural advantages of skeletons to obtain the robustness to cloth changing for gait recognition.
%
%---Horizontal OUMVLP Experiments---
\begin{table}[!t]
\newcommand{\bftab}[1]{{\fontseries{b}\selectfont#1}}
\caption{The rank-1 accuracy (\%) on OUMVLP across different views, excluding the identical-view cases\label{table_oumvlp}}
\begin{center}
\resizebox{0.99\textwidth}{0.09\textheight}{
\begin{tabular}{ccccccccccccccccc}
\toprule
\multicolumn{1}{c}{Gallery \#01, Probe \#00} & 
\multicolumn{14}{c}{Probe View} & 
\multirow{2}{*}{mean} \\ \cline{1-15}

\multicolumn{1}{c}{Methods} &
 $0^{\circ}$ & $15^{\circ}$ & $30^{\circ}$ & $45^{\circ}$ & $60^{\circ}$ & $75^{\circ}$ & $90^{\circ}$ & $180^{\circ}$ & $195^{\circ}$ & $210^{\circ}$ & $225^{\circ}$ & $240^{\circ}$ &
 $255^{\circ}$ &
\multicolumn{1}{l}{$270^{\circ}$} &\\ \cline{1-16}

\multicolumn{1}{c}{ST-GCN~\cite{sijieyan2018st-gcn}} & 
24.3 &34.9 &39.8 &42.1 &41.5 &38.5 &33.3 &23.0 &27.0 &25.7 &35.6 &35.3 &32.5 &
\multicolumn{1}{l}{28.2} & \multicolumn{1}{c}{33.0}\\ 

\multicolumn{1}{c}{MSGG(ours)} & 
43.8 &58.8 &64.0 &66.4 &65.9 &62.9 &57.8 &40.6 &48.4 &44.4 &60.6 &60.3 &56.6 &
\multicolumn{1}{l}{51.8} & \multicolumn{1}{c}{55.9}\\ \cline{1-16}

\multicolumn{1}{c}{GaitSet~\cite{chao2019gaitset}} &
79.5 &87.9 &89.9 &90.2 &88.1 &88.7 &87.8 &81.7 &86.7 &89.0 &89.3 &87.2 &87.8 &
\multicolumn{1}{c}{86.2} & \multicolumn{1}{c}{87.1}\\ 

\multicolumn{1}{c}{GaitPart~\cite{chaofan2020gaitpart}} &
82.6 &88.9 &90.8 &91.0 &89.7 &89.9 &89.5 &85.2 &88.1 &90.0 &90.1 &89.0 &89.1 &
\multicolumn{1}{c}{88.2} & \multicolumn{1}{c}{88.7}\\ 

\multicolumn{1}{c}{GLN~\cite{hou2020gait}} &
83.8 &90.0 &91.0 &91.2 &90.3 &90.0 &89.4 &85.3 &89.1 &\bftab{90.5} &90.6 &89.6 &89.3 & 
\multicolumn{1}{c}{88.5} & \multicolumn{1}{c}{89.2}\\

\multicolumn{1}{c}{GaitGL~\cite{lin2021gaitgl}} &
84.9 &90.2 &91.1 &91.5 &\bftab{91.1} &\bftab{90.8} &90.3 &\bftab{88.5} &88.6 &90.3 &90.4 &89.6 &89.5 &
\multicolumn{1}{c}{88.8} & \multicolumn{1}{c}{89.7}\\ \cline{1-16}

\multicolumn{1}{c}{silhouette-module(base)} &
82.57 &88.93 &90.84 &91.00 &89.75 &89.91 &89.50 &85.19 &88.09 &90.02 &90.15 &89.03 &89.10 &
\multicolumn{1}{c}{88.24} & \multicolumn{1}{c}{88.74}\\

\multicolumn{1}{c}{BiFusion(ours)} & 
\bftab{86.17} &\bftab{90.60} &\bftab{91.28} &\bftab{91.56} &90.88 &\bftab{90.78} &\bftab{90.48} &87.76 &\bftab{89.48} &90.38 &\bftab{90.65} &\bftab{89.95} &\bftab{89.82} &
\multicolumn{1}{l}{\bftab{89.32}} & \multicolumn{1}{c}{\bftab{89.94}}\\ \bottomrule
\end{tabular}
}
\end{center}
\end{table}

\noindent\textbf{OUMVLP}
The performance comparison on OUMVLP is shown in Table \ref{table_oumvlp}.
Since the division provided in OUMVLP~\cite{takemura2018oumvlp} is not consistent with that in OUMVLP-Pose~\cite{weizhian2020oumvlp-pose}, we conduct the experiment on ST-GCN (baseline) for the comparison with the proposed MSGG network.
As shown in Table \ref{table_oumvlp}, the MSGG achieves better results under all the probe views, which demonstrates its effectiveness for the skeleton-based gait recognition.

We have noticed that there is a huge performance gap of the proposed MSGG performed on CASIA-B and OUMVLP-Pose.
The possible reasons for this phenomenon can be summarized into two aspects.
Firstly, considering the huge gaps in the rank-1 accuracies of the proposed MSGG performed on the skeleton data predicted by HRNet~\cite{kesun2019pose} compared with that on the skeleton data predicted by Openpose~\cite{cao2017openpose} (i.e., NM+8.8\%, BG+11.5\%, and CL+12.4\% in Table \ref{table_casia-b}), the quality of the predicted skeleton data greatly influence the final recognition results.
The OUMVLP-Pose used in Table~\ref{table_oumvlp} is predicted by AlphaPose (OUMVLP-Pose provides the skeleton data predicted by AlphaPose~\cite{fang2017alphapose} and Openpose~\cite{cao2017openpose}; the reason to choose the skeleton data of AlphaPose is due to its better recognition results for both ST-GCN and MSGG compared with that of Openpose).
The poor quality of the skeleton data in OUMVLP-Pose may cause the poor performance of the proposed MSGG in Table~\ref{table_oumvlp}. 
Secondly, the small number of frames (roughly averaged at 25) in a skeleton sequence in OUMVLP-Pose may lead to poor performance too.
In order to retain most of the data during training, the frame number of a skeleton sequence in a batch is set to 18, which is less than the frame number of a complete gait cycle.
Furthermore, the above phenomenon is consistent with the performance gap of the ST-GCN~\cite{sijieyan2018st-gcn} performed on CASIA-B and OUMVLP-Pose (i.e., NM-85.4\%, BG-64.2\%, and 59.5\% for the ST-GCN on CASIA-B and 33.0\% for the ST-GCN on OUMVLP-Pose).

Furthermore, the proposed BiFusion network improves the mean accuracy on its silhouette module by 1.2\%. 
This is less compared to the improvements in CASIA-B.
Considering that the major improvements in CASIA-B are made on CL, the OUMVLP only contains walkings in normal and this may lead to less improvement.
Moreover, the poor performance of the AlphaPose~\cite{fang2017alphapose} and the small number of frames of a sequence (roughly averaged at 25) in OUMVLP-Pose may bring negative effects.

\subsection{Comparison with the Collaborative Network}
In this section, we provide a performance comparison with the Collaborative Network~\cite{yao2021gait} which also combines silhouettes and skeletons but uses additional RGB images and utilizes the skeleton data differently.
%
% In the Collaborative Network, additional RGB images are used and the exploitation of the skeleton data is different from the proposed BiFusion network.
%
Specifically, the Collaborative Network uses the skeleton to locate local features on RGB frames while the proposed BiFusion network fully exploits the discriminative gait patterns contained in skeletons.
The part-based silhouette features are concatenated to the skeleton-based RGB features in the Collaborative Network and the multi-scale skeleton features in the BiFusion network, respectively.
The Collaborative Network conducts experiments on the CASIA-B dataset and provides only the accuracies of CL and BG since it is intended for solving the cloth-changing problem.
Under the setting of training with the first 74 subjects and testing with the last 50 subjects, the accuracies of CL and BG for the Collaborative Network are 95.3\% and 90.6\%, which are lower than that of the proposed BiFusion network (BG-96.0\% and CL-92.1\%).
It is worth noting that the accuracies are up to NM-100\%, BG-98.8\%, and CL-91.4\% on CASIA-B in CHD~\cite{cai2019chd} which takes RGB images as the input and performs the person re-identification task~\cite{zheng2016personreid}.
The Collaborative Network uses estimated keypoint heat maps to extract local features on RGB frames directly while the proposed BiFusion network only uses skeletons and silhouettes.
In spite of this, the better performance of the proposed BiFusion network compared with the Collaborative Network demonstrates the benefits of simultaneously exploiting gait patterns from skeletons and silhouettes.
%
% %---Vertical Ablation Study---
% \begin{table}[t]
% \begin{center}
% \caption{Ablation experiments conducted on CASIA-B for the proposed MSGG. Results are reported in the rank-1 accuracy (\%) averaged on all 11 views, excluding the identical-view cases.\label{table_ablation_study_msgg}}
% \resizebox{0.489\textwidth}{18.5mm}{
% \begin{tabular}{c|c|c|c|c}
% \toprule
% \multirow{2}{*}{Partition Strategy} & \multirow{2}{*}{Pyramid Structure} & \multicolumn{3}{c}{Probes} \\ \cline{3-5}
% & &\multicolumn{1}{c|}{NM} &
% \multicolumn{1}{c|}{BG} &
% \multicolumn{1}{c}{CL} \\ \hline
% Uni-labeling &Joints+Limbs+Bodyparts &92.7 &77.8 &64.3\\ \hline
% Distance &Joints+Limbs+Bodyparts &92.1 &74.4 &63.0\\ \hline
% Spatial &Joints+Limbs+Bodyparts &90.0 &71.7 &62.7\\ \hline
% Gait Temporal &Joints+Limbs+Bodyparts &\textbf{93.0} &\textbf{78.1} &\textbf{68.3}\\ \hline
% Gait Temporal &Joints+Limbs &91.1 &75.1 &66.2\\ \hline
% Gait Temporal &Joints &89.2 &72.2 &\textbf{68.4}\\ \hline
% Gait Temporal &Joints+Joints+Joints &82.8 &63.0 &59.0\\ \hline
% Gait Temporal &Joints+Limbs+Bodyparts (Separate) &90.9 &72.8 &58.6\\ \bottomrule
% \end{tabular}
% }
% \end{center}
% \end{table}
%

%---Vertical Ablation Study---
\begin{table}[!t]
\caption{Ablation experiments conducted on CASIA-B for the MSGG. Results are reported in the rank-1 accuracy (\%) averaged on all 11 views, excluding the identical-view cases}
\begin{center}
\resizebox{0.68\textwidth}{0.09\textheight}{
\begin{tabular}{ccccc}
\toprule
\multirow{2}{*}{Partition Strategy} & \multirow{2}{*}{Pyramid Structure} & \multicolumn{3}{c}{Probes} \\ \cline{3-5}
& &\multicolumn{1}{c}{NM} &
\multicolumn{1}{c}{BG} &
\multicolumn{1}{c}{CL} \\ \hline
Uni-labeling &Joints+Limbs+Bodyparts &92.7 &77.8 &64.3\\ \hline
Distance &Joints+Limbs+Bodyparts &92.1 &74.4 &63.0\\ \hline
Spatial &Joints+Limbs+Bodyparts &90.0 &71.7 &62.7\\ \hline
Gait Temporal &Joints+Limbs+Bodyparts &\textbf{93.0} &\textbf{78.1} &\textbf{68.3}\\ \hline
Gait Temporal &Joints+Limbs &91.1 &75.1 &66.2\\ \hline
Gait Temporal &Joints &89.2 &72.2 &\textbf{68.4}\\ \hline
Gait Temporal &Joints+Joints+Joints &82.8 &63.0 &59.0\\ \hline
Gait Temporal &Joints+Limbs+Bodyparts (Separate) &90.9 &72.8 &58.6\\ \bottomrule
\end{tabular}
}
\end{center}
\label{table_ablation_study_msgg}
\end{table}

\subsection{Ablation Study}
\label{ablation_msgg}
In this section, we conduct the ablation study to analyze the behaviors of MSGG and BiFusion. 
The experiments are mainly conducted on CASIA-B which contains the skeleton data of high quality (estimated by HRNet) and covers different walking conditions (NM-normal walking, BG-walking with bags, CL-walking in different clothes).
The experimental settings are basically the same as those in Sec.~\ref{table_ablation_study_msgg} and we report the rank-1 accuracy averaged on all views excluding the identical-view cases for simplicity.

\textbf{Impact of the node partition strategy}
The first 4 rows in Table \ref{table_ablation_study_msgg} compare the performance of different node partition strategies, where three of them are directly taken from the ST-GCN~\cite{sijieyan2018st-gcn}.
The proposed \emph{Gait temporal partitioning} strategy improves the accuracy on CASIA-B by at least +0.3\% for NM, +0.3\% for BG, and +4.0\% for CL. 
\begin{figure}  
\centering  
\includegraphics[width=0.99\textwidth]{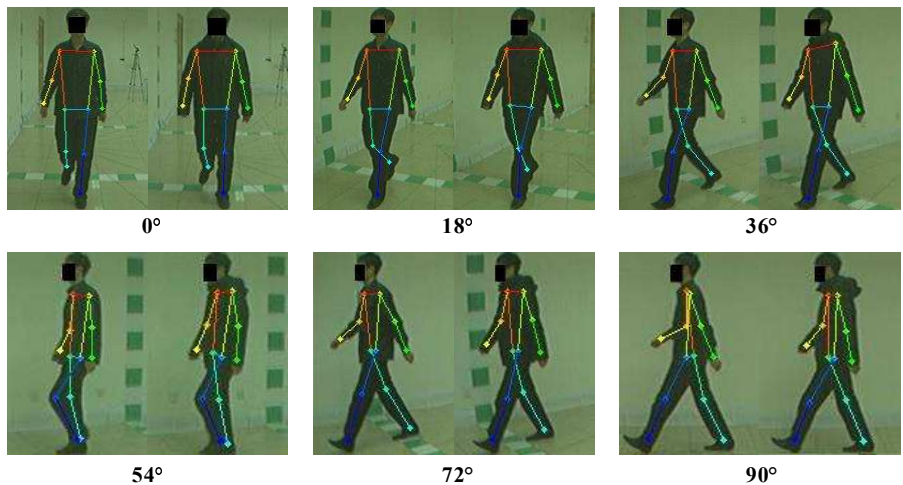}
\caption{The visualization of pose estimation results for the same person across different clothes under the view of $0^{\circ}$ to $90^{\circ}$, respectively. The left side of each pair of pictures belongs to normal walking (NM) and the right side belongs to walking with different clothes (CL)}\label{fig_pose_across_clothes}
\end{figure}

\textbf{Impact of the pyramid structure}
The last 5 rows in Table \ref{table_ablation_study_msgg} show the effectiveness of the proposed pyramid structure.
In MSGG, there are three branches conducted on three spatial-temporal subgraphs respectively, corresponding to three different scales.
The experimental results of the first branch (Joints), the first two branches (Joints+Limbs), and the whole MSGG (Joints+Limbs+Bodyparts) are listed in lines 6, 5, and 4 respectively.
It can be seen that the accuracies are constantly improved with the increasing number of branches for both NM (+3.8\%) and BG (+5.9\%).
We have noticed that the accuracies of CL for the first branch (Joints) and the whole MSGG (Joints+Limbs+Bodyparts) are comparable.
The accuracy of CL is calculated under the conditions of taking normal walkings as the gallery and taking walkings with different clothes as the probe.
The original differences of predicted skeletons across clothes (shown in Fig.~\ref{fig_pose_across_clothes}) are one possible reason for this phenomenon.
Since the multi-scale architecture design may bring error accumulation and could not handle this problem well, the gap between estimated skeletons across clothes needs further exploration.

Secondly, the results in the $7^{th}$ row are the performance of the network with three joints branches.
Except that the three branches are conducted on the joints spatial-temporal subgraph, the experimental settings of the $7^{th}$ row are consistent with that of the $4^{th}$ row.
Therefore, the comparison of the $4^{th}$ row and the $7^{th}$ row demonstrates the effectiveness of the proposed multi-scale skeleton structure. 
Moreover, we conduct the experiment that turns off the message passing across different scales, i.e., removes the Semantic Pooling operation in the Cross-scale Spatial-Temporal Block, and concatenates the features of different scales (branches) for recognition.
Comparing the results in the $4^{th}$ row with that in the last row, the cross-scale message passing in MSGG improves the rank-1 accuracy by +2.1\% for NM, +5.3\% for BG, and +9.7\% for CL.

\textbf{Impact of the bimodal fusion}
The third block of different probes in Table \ref{table_casia-b} compares the mean accuracies of the BiFusion with that of its silhouette module (GaitPart).
The effectiveness of utilizing the complementary strengths in different modalities for gait recognition is empirically studied.
Compared with the silhouette-based GaitPart module, the Bimodal Fusion network improves the rank-1 accuracy by +2.6\% for NM, +4.6\% for BG, and +12.6\% for CL.

\section{Conclusion}
In this work, the inherent hierarchical semantics of body joints in a skeleton is leveraged to design a novel Multi-Scale Gait Graph (MSGG) network for capturing intrinsic discrimination of the skeleton data.
Different from previous multimodal methods that just employ the skeleton as a locator to assist the local feature extraction on silhouettes or RGB frames, we propose a simple yet effective Bimodal Fusion network that integrates crucial discriminative information retained in both the silhouette and the skeleton for gait identification.
Extensive experiments on CASIA-B and OUMVLP demonstrate both the superiority of the proposed MSGG network among skeleton-based methods and the effectiveness of the bimodal fusion for gait recognition.

The flexibility of the GCN-based gait recognition opens up many possible directions for future works on skeleton data, e.g., automatical node partitioning and hierarchy constructing.
Furthermore, the combination of features extracted from skeletons and silhouettes shows significant improvements in recognizing gait.
It is expected to open up a new research hotspot of mining intrinsic discrimination of different data modalities for gait identification.
The fusion of different data modalities and the way to reach a more comprehensive representation of gait remain to be further studied.

\section*{Declarations}
\subsection*{Funding}
No funding was received to assist with the preparation of this manuscript.
\subsection*{Conflict of interest}
The authors have no relevant financial or non-financial interests to disclose.
\subsection*{Data availability}
The data that support the findings of this study are available on request from the Institute of Automation, Chinese Academy of Sciences (CASIA) (http://www.cbsr.ia.ac.cn/english/Gait\%20Databases.asp) and the Institute of Scientific and Industrial Research (ISIR), Osaka University (OU) (http://www.am.sanken.osaka-u.ac.jp/BiometricDB/GaitMVLP.html).

\bibliography{manuscript}% common bib file

\begin{thebibliography}{56}
\providecommand{\natexlab}[1]{#1}
\providecommand{\url}[1]{{#1}}
\providecommand{\urlprefix}{URL }
\providecommand{\doi}[1]{\url{https://doi.org/#1}}
\providecommand{\eprint}[2][]{\url{#2}}
 \bibcommenthead

\bibitem[{Zhang et~al(2019)Zhang, Tran, Yin, Atoum, Liu, Wan, and
  Wang}]{ziyuanzhang2019gait}
Zhang Z, Tran L, Yin X, et~al (2019) Gait recognition via disentangled
  representation learning. In: Proceedings of the IEEE/CVF Conference on
  Computer Vision and Pattern Recognition, pp 4710--4719,
  \doi{10.1109/cvpr.2019.00484}

\bibitem[{Sun et~al(2014)Sun, Wang, and Tang}]{yisun2014face}
Sun Y, Wang X, Tang X (2014) Deep learning face representation by joint
  identification-verification. Advances in neural information processing
  systems 27

\bibitem[{Maltoni et~al(2005)Maltoni, Maio, Jain, and
  Prabhakar}]{maltoni2005fingerprint}
Maltoni D, Maio D, Jain A, et~al (2005) Handbook of fingerprint recognition. Ch
  Synthetic Fingerprint Generation 33(5-6):1314

\bibitem[{Wildes and R.P.(1997)}]{wildes1997Iris}
Wildes, R.P. (1997) Iris recognition: an emerging biometric technology.
  Proceedings of the IEEE 85(9):1348--1363. \doi{10.1109/5.628669}

\bibitem[{Larsen et~al(2008)Larsen, Simonsen, and Lynnerup}]{larsen2008gait}
Larsen PK, Simonsen EB, Lynnerup N (2008) Gait analysis in forensic medicine.
  Journal of Forensic Sciences 53(5):1149--1153.
  \doi{10.1111/j.1556-4029.2008.00807.x}

\bibitem[{Bouchrika et~al(2011)Bouchrika, Goffredo, Carter, and
  Nixon}]{bouchrika2011gait}
Bouchrika I, Goffredo M, Carter J, et~al (2011) On using gait in forensic
  biometrics. Journal of forensic sciences 56(4):882--889.
  \doi{10.1111/j.1556-4029.2011.01793.x}

\bibitem[{Zhang et~al(2020)Zhang, Tran, Liu, and Liu}]{ziyuanzhang2020gait}
Zhang Z, Tran L, Liu F, et~al (2020) On learning disentangled representations
  for gait recognition. IEEE Transactions on Pattern Analysis and Machine
  Intelligence \doi{10.1109/tpami.2020.2998790}

\bibitem[{Li et~al(2020)Li, Makihara, Xu, Yagi, Yu, and Ren}]{li2020gait}
Li X, Makihara Y, Xu C, et~al (2020) End-to-end model-based gait recognition.
  In: Proceedings of the Asian Conference on Computer Vision, pp 3--20

\bibitem[{Wang et~al(2021)Wang, Zhang, Shen, Du, Zhao, Lizhen, and
  Wen}]{wang2021eventgait}
Wang Y, Zhang X, Shen Y, et~al (2021) Event-stream representation for human
  gaits identification using deep neural networks. IEEE Transactions on Pattern
  Analysis and Machine Intelligence \doi{10.1109/tpami.2021.3054886}

\bibitem[{Gallego et~al(2020)Gallego, Delbruck, Orchard, Bartolozzi, Taba,
  Censi, Leutenegger, Davison, Conradt, Daniilidis et~al}]{gallego2020event}
Gallego G, Delbruck T, Orchard GM, et~al (2020) Event-based vision: A survey.
  IEEE Transactions on Pattern Analysis and Machine Intelligence
  \doi{10.1109/tpami.2020.3008413}

\bibitem[{Chao et~al(2019)Chao, He, Zhang, and Feng}]{chao2019gaitset}
Chao H, He Y, Zhang J, et~al (2019) Gaitset: Regarding gait as a set for
  cross-view gait recognition. In: Proceedings of the AAAI conference on
  artificial intelligence, pp 8126--8133, \doi{10.1609/aaai.v33i01.33018126}

\bibitem[{{Fan} et~al(2020){Fan}, {Peng}, {Cao}, {Liu}, {Hou}, {Chi}, {Huang},
  {Li}, and {He}}]{chaofan2020gaitpart}
{Fan} C, {Peng} Y, {Cao} C, et~al (2020) Gaitpart: Temporal part-based model
  for gait recognition. In: 2020 IEEE/CVF Conference on Computer Vision and
  Pattern Recognition (CVPR), pp 14,225--14,233,
  \doi{10.1109/cvpr42600.2020.01423}

\bibitem[{Lin et~al(2021)Lin, Zhang, and Yu}]{lin2021gaitgl}
Lin B, Zhang S, Yu X (2021) Gait recognition via effective global-local feature
  representation and local temporal aggregation. In: Proceedings of the
  IEEE/CVF International Conference on Computer Vision, pp 14,648--14,656,
  \doi{10.1109/iccv48922.2021.01438}

\bibitem[{Hou et~al(2020)Hou, Cao, Liu, and Huang}]{hou2020gait}
Hou S, Cao C, Liu X, et~al (2020) Gait lateral network: Learning discriminative
  and compact representations for gait recognition. In: European Conference on
  Computer Vision, Springer, pp 382--398, \doi{10.1007/978-3-030-58545-7_22}

\bibitem[{Ding et~al(2021)Ding, Wang, Wang, Lan, and Liu}]{xinnan2021gait}
Ding X, Wang K, Wang C, et~al (2021) Sequential convolutional network for
  behavioral pattern extraction in gait recognition. Neurocomputing
  463:411--421. \doi{10.1016/j.neucom.2021.08.054}

\bibitem[{Sun et~al(2018)Sun, Wang, Li, Wan, Cheng, and Zhang}]{sun2018view}
Sun J, Wang Y, Li J, et~al (2018) View-invariant gait recognition based on
  kinect skeleton feature. Multimedia Tools and Applications
  77(19):24,909--24,935. \doi{10.1007/s11042-018-5722-1}

\bibitem[{Liao et~al(2020)Liao, Yu, An, and Huang}]{liao2020gait}
Liao R, Yu S, An W, et~al (2020) A model-based gait recognition method with
  body pose and human prior knowledge. Pattern Recognition 98:107,069.
  \doi{10.1016/j.patcog.2019.107069}

\bibitem[{An et~al(2020)An, Yu, Makihara, Wu, Xu, Yu, Liao, and
  Yagi}]{weizhian2020oumvlp-pose}
An W, Yu S, Makihara Y, et~al (2020) Performance evaluation of model-based gait
  on multi-view very large population database with pose sequences. IEEE
  Transactions on Biometrics, Behavior, and Identity Science 2(4):421--430.
  \doi{10.1109/tbiom.2020.3008862}

\bibitem[{Mao and Song(2020)}]{mao2020gait}
Mao M, Song Y (2020) Gait recognition based on 3d skeleton data and graph
  convolutional network. In: 2020 IEEE International Joint Conference on
  Biometrics (IJCB), \doi{10.1109/ijcb48548.2020.9304916}

\bibitem[{Teepe et~al(2021)Teepe, Khan, Gilg, Herzog, Hörmann, and
  Rigoll}]{teepe2021gaitgraph}
Teepe T, Khan A, Gilg J, et~al (2021) Gaitgraph: Graph convolutional network
  for skeleton-based gait recognition. In: 2021 IEEE International Conference
  on Image Processing (ICIP). IEEE, pp 2314--2318,
  \doi{10.1109/icip42928.2021.9506717}

\bibitem[{Xu et~al(2021)Xu, Jiang, and Sun}]{xu2021gait}
Xu K, Jiang X, Sun T (2021) Gait identification based on human skeleton with
  pairwise graph convolutional network. In: 2021 IEEE International Conference
  on Multimedia and Expo (ICME), IEEE, pp 1--6,
  \doi{10.1109/icme51207.2021.9428123}

\bibitem[{Cao et~al(2017)Cao, Simon, Wei, and Sheikh}]{cao2017openpose}
Cao Z, Simon T, Wei SE, et~al (2017) Realtime multi-person 2d pose estimation
  using part affinity fields. In: 2017 IEEE Conference on Computer Vision and
  Pattern Recognition (CVPR), IEEE Computer Society, pp 1302--1310,
  \doi{10.1109/cvpr.2017.143}

\bibitem[{Sun et~al(2019)Sun, Xiao, Liu, and Wang}]{kesun2019pose}
Sun K, Xiao B, Liu D, et~al (2019) Deep high-resolution representation learning
  for human pose estimation. In: Proceedings of the IEEE/CVF Conference on
  Computer Vision and Pattern Recognition (CVPR), \doi{10.1109/cvpr.2019.00584}

\bibitem[{{Boulgouris} and {Huang}(2013)}]{boulgouris2013gait}
{Boulgouris} NV, {Huang} X (2013) Gait recognition using hmms and dual
  discriminative observations for sub-dynamics analysis. IEEE Transactions on
  Image Processing 22(9):3636--3647. \doi{10.1109/tip.2013.2266578}

\bibitem[{Yao et~al(2021)Yao, Kusakunniran, Wu, Xu, and Zhang}]{yao2021gait}
Yao L, Kusakunniran W, Wu Q, et~al (2021) Collaborative feature learning for
  gait recognition under cloth changes. IEEE Transactions on Circuits and
  Systems for Video Technology \doi{10.1109/tcsvt.2021.3112564}

\bibitem[{Han and Bhanu(2005)}]{hanjun2005gait}
Han J, Bhanu B (2005) Individual recognition using gait energy image. IEEE
  Transactions on Pattern Analysis and Machine Intelligence 28(2):316--322.
  \doi{10.1109/tpami.2006.38}

\bibitem[{{Yu} et~al(2017){Yu}, {Chen}, {Wang}, {Shen}, and
  {Huang}}]{yu2017gait}
{Yu} S, {Chen} H, {Wang} Q, et~al (2017) Invariant feature extraction for gait
  recognition using only one uniform model. Neurocomputing 239:81--93.
  \doi{10.1016/j.neucom.2017.02.006}

\bibitem[{Aggarwal and Vishwakarma(2018)}]{himanshu2018gait}
Aggarwal H, Vishwakarma DK (2018) Covariate conscious approach for gait
  recognition based upon zernike moment invariants. IEEE Transactions on
  Cognitive and Developmental Systems 10(2):397--407.
  \doi{10.1109/tcds.2017.2658674}

\bibitem[{Lishani et~al(2019)Lishani, Boubchir, Khalifa, and
  Bouridane}]{lishani2019gait}
Lishani AO, Boubchir L, Khalifa E, et~al (2019) Human gait recognition using
  gei-based local multi-scale feature descriptors. Multimedia Tools and
  Applications 78(5):5715--5730. \doi{10.1007/s11042-018-5752-8}

\bibitem[{Xu et~al(2019)Xu, Makihara, Li, Yagi, and Lu}]{xu2019gait}
Xu C, Makihara Y, Li X, et~al (2019) Speed-invariant gait recognition using
  single-support gait energy image. Multimedia Tools and Applications
  78(18):26,509--26,536. \doi{10.1007/s11042-019-7712-3}

\bibitem[{Tong et~al(2018)Tong, Fu, Yue, and Ling}]{suibing2018gait}
Tong S, Fu Y, Yue X, et~al (2018) Multi-view gait recognition based on a
  spatial-temporal deep neural network. IEEE Access 6:57,583--57,596.
  \doi{10.1109/access.2018.2874073}

\bibitem[{Lin et~al(2020)Lin, Zhang, and Bao}]{lin2020gait}
Lin B, Zhang S, Bao F (2020) Gait recognition with multiple-temporal-scale 3d
  convolutional neural network. In: Proceedings of the 28th ACM International
  Conference on Multimedia, pp 3054--3062, \doi{10.1145/3394171.3413861}

\bibitem[{Deng and Wang(2018)}]{deng2018gait}
Deng M, Wang C (2018) Human gait recognition based on deterministic learning
  and data stream of microsoft kinect. IEEE Transactions on Circuits and
  Systems for Video Technology 29(12):3636--3645.
  \doi{10.1109/tcsvt.2018.2883449}

\bibitem[{Yan et~al(2018)Yan, Xiong, and Lin}]{sijieyan2018st-gcn}
Yan S, Xiong Y, Lin D (2018) Spatial temporal graph convolutional networks for
  skeleton-based action recognition. In: Thirty-second AAAI conference on
  artificial intelligence

\bibitem[{Liu et~al(2021)Liu, Zha, Wu, Zheng, and Sun}]{liu2021videoreid}
Liu J, Zha ZJ, Wu W, et~al (2021) Spatial-temporal correlation and topology
  learning for person re-identification in videos. In: Proceedings of the
  IEEE/CVF Conference on Computer Vision and Pattern Recognition, pp
  4370--4379, \doi{10.1109/cvpr46437.2021.00435}

\bibitem[{Bodla et~al(2017)Bodla, Zheng, Xu, Chen, Castillo, and
  Chellappa}]{bodla2017fusion}
Bodla N, Zheng J, Xu H, et~al (2017) Deep heterogeneous feature fusion for
  template-based face recognition. In: 2017 IEEE winter conference on
  applications of computer vision (WACV), IEEE, pp 586--595,
  \doi{10.1109/wacv.2017.71}

\bibitem[{Xin et~al(2018)Xin, Kong, Liu, Wang, Zhu, Gao, Zhao, and
  Xu}]{xin2018fusion}
Xin Y, Kong L, Liu Z, et~al (2018) Multimodal feature-level fusion for
  biometrics identification system on iomt platform. IEEE Access pp 1--1.
  \doi{10.1109/access.2018.2815540}

\bibitem[{Dhiman and Vishwakarma(2020)}]{dhiman2020multimodalaction}
Dhiman C, Vishwakarma DK (2020) View-invariant deep architecture for human
  action recognition using two-stream motion and shape temporal dynamics. IEEE
  Transactions on Image Processing 29:3835--3844.
  \doi{10.1109/tip.2020.2965299}

\bibitem[{Singh and
  Vishwakarma(2021{\natexlab{a}})}]{singh2021multimodalaction}
Singh T, Vishwakarma DK (2021{\natexlab{a}}) A deeply coupled convnet for human
  activity recognition using dynamic and rgb images. Neural Computing and
  Applications 33(1):469--485. \doi{10.1007/s00521-020-05018-y}

\bibitem[{Singh and
  Vishwakarma(2021{\natexlab{b}})}]{singh2021multimodalaction2}
Singh T, Vishwakarma DK (2021{\natexlab{b}}) A deep multimodal network based on
  bottleneck layer features fusion for action recognition. Multimedia Tools and
  Applications 80(24):33,505--33,525. \doi{10.1007/s11042-021-11415-9}

\bibitem[{Dhiman et~al(2021)Dhiman, Vishwakarma, and
  Agarwal}]{dhiman2021multimodalaction}
Dhiman C, Vishwakarma DK, Agarwal P (2021) Part-wise spatio-temporal attention
  driven cnn-based 3d human action recognition. ACM Transactions on Multimidia
  Computing Communications and Applications 17(3):1--24. \doi{10.1145/3441628}

\bibitem[{Ross and Govindarajan(2005)}]{ross2005fusion}
Ross AA, Govindarajan R (2005) Feature level fusion of hand and face
  biometrics. In: Biometric technology for human identification II,
  International Society for Optics and Photonics, pp 196--204,
  \doi{10.1117/12.606093}

\bibitem[{{Faundez-Zanuy}(2005)}]{faundez-zanuy2005fusion}
{Faundez-Zanuy} M (2005) Data fusion in biometrics. IEEE Aerospace and
  Electronic Systems Magazine 20(1):34--38. \doi{10.1109/maes.2005.1396793}

\bibitem[{{Shekhar} et~al(2014){Shekhar}, {Patel}, {Nasrabadi}, and
  {Chellappa}}]{shekhar2014fusion}
{Shekhar} S, {Patel} VM, {Nasrabadi} NM, et~al (2014) Joint sparse
  representation for robust multimodal biometrics recognition. IEEE
  Transactions on Pattern Analysis and Machine Intelligence 36(1):113--126.
  \doi{10.1109/tpami.2013.109}

\bibitem[{Iwama et~al(2012)Iwama, Okumura, Makihara, and Yagi}]{iwama2012isir}
Iwama H, Okumura M, Makihara Y, et~al (2012) The ou-isir gait database
  comprising the large population dataset and performance evaluation of gait
  recognition. IEEE Transactions on Information Forensics and Security
  7(5):1511--1521. \doi{10.1109/tifs.2012.2204253}

\bibitem[{Makihara et~al(2012)Makihara, Mannami, Tsuji, Hossain, Sugiura, Mori,
  and Yagi}]{makihara2012isirtreadmill}
Makihara Y, Mannami H, Tsuji A, et~al (2012) The ou-isir gait database
  comprising the treadmill dataset. IPSJ Transactions on Computer Vision and
  Applications 4:53--62. \doi{10.2197/ipsjtcva.4.53}

\bibitem[{Yu et~al(2006)Yu, Tan, and Tan}]{shiqiyu2006gait}
Yu S, Tan D, Tan T (2006) A framework for evaluating the effect of view angle,
  clothing and carrying condition on gait recognition. In: 18th International
  Conference on Pattern Recognition (ICPR 2006), 20-24 August 2006, Hong Kong,
  China, \doi{10.1109/icpr.2006.67}

\bibitem[{Takemura et~al(2018)Takemura, Makihara, Muramatsu, Echigo, and
  Yagi}]{takemura2018oumvlp}
Takemura N, Makihara Y, Muramatsu D, et~al (2018) Multi-view large population
  gait dataset and its performance evaluation for cross-view gait recognition.
  IPSJ Transactions on Computer Vision and Applications 10(1):1--14.
  \doi{10.1186/s41074-018-0039-6}

\bibitem[{Fang et~al(2017)Fang, Xie, Tai, and Lu}]{fang2017alphapose}
Fang HS, Xie S, Tai YW, et~al (2017) Rmpe: Regional multi-person pose
  estimation. In: Proceedings of the IEEE international conference on computer
  vision, pp 2334--2343, \doi{10.1109/iccv.2017.256}

\bibitem[{{Paszke} et~al(2019){Paszke}, {Gross}, {Massa}, {Lerer}, {Bradbury},
  {Chanan}, {Killeen}, {Lin}, {Gimelshein}, {Antiga}, {Desmaison}, {Kopf},
  {Yang}, {DeVito}, {Raison}, {Tejani}, {Chilamkurthy}, {Steiner}, {Fang},
  {Bai}, and {Chintala}}]{paszke2019pytorch}
{Paszke} A, {Gross} S, {Massa} F, et~al (2019) Pytorch: An imperative style,
  high-performance deep learning library. In: Advances in Neural Information
  Processing Systems, pp 8026--8037

\bibitem[{Hermans et~al(2017)Hermans, Beyer, and
  Leibe}]{hermans2017tripletloss}
Hermans A, Beyer L, Leibe B (2017) In defense of the triplet loss for person
  re-identification. \eprint{1703.07737}

\bibitem[{Chen and Ramanan(2017)}]{chen20173dpose}
Chen C, Ramanan D (2017) 3d human pose estimation = 2d pose estimation +
  matching. In: 2017 IEEE Conference on Computer Vision and Pattern Recognition
  (CVPR). IEEE Computer Society, pp 5759--5767, \doi{10.1109/cvpr.2017.610}

\bibitem[{Wu et~al(2017)Wu, Huang, Wang, Wang, and Tan}]{zifengwu2017gait}
Wu Z, Huang Y, Wang L, et~al (2017) A comprehensive study on cross-view gait
  based human identification with deep cnns. IEEE Transactions on Pattern
  Analysis \& Machine Intelligence 39(02):209--226.
  \doi{10.1109/tpami.2016.2545669}

\bibitem[{Chen et~al(2021)Chen, Luo, Weng, Luo, Li, and Tian}]{xinchen2021gait}
Chen X, Luo X, Weng J, et~al (2021) Multi-view gait image generation for
  cross-view gait recognition. IEEE Transactions on Image Processing
  30:3041--3055. \doi{10.1109/tip.2021.3055936}

\bibitem[{Cai et~al(2019)Cai, Zhou, and Wang}]{cai2019chd}
Cai C, Zhou Y, Wang Y (2019) Chd: Consecutive horizontal dropout for human gait
  feature extraction. In: Proceedings of the 2019 8th International Conference
  on Computing and Pattern Recognition, pp 89--94,
  \doi{10.1145/3373509.3373556}

\bibitem[{Zheng et~al(2016)Zheng, Yang, and Hauptmann}]{zheng2016personreid}
Zheng L, Yang Y, Hauptmann AG (2016) Person re-identification: Past, present
  and future. arXiv preprint arXiv:161002984

\end{thebibliography}

\end{document}